# A Note On The Popularity of Stochastic Optimization Algorithms in Different Fields: A Quantitative Analysis from 2007 to 2017

Son Duy Dao

IMT Atlantique, Brest, France

Email: son-duy.dao@imt-atlantique.fr

**Abstract:** Stochastic optimization algorithms are often used to solve complex large-scale optimization problems in various fields. To date, there have been a number of stochastic optimization algorithms such as Genetic Algorithm, Cuckoo Search, Tabu Search, Simulated Annealing, Particle Swarm Optimization, Ant Colony Optimization, etc. Each algorithm has some advantages and disadvantages. Currently, there is no study that can help researchers to choose the most popular optimization algorithm to deal with the problems in different research fields. In this note, a quantitative analysis of the popularity of 14 stochastic optimization algorithms in 18 different research fields in the last ten years from 2007 to 2017 is provided. This quantitative analysis can help researchers/practitioners select the best optimization algorithm to solve complex large-scale optimization problems in the fields of Engineering, Computer science, Operations research, Mathematics, Physics, Chemistry, Automation control systems, Materials science, Energy fuels, Mechanics, Telecommunications, Thermodynamics, Optics, Environmental sciences ecology, Water resources, Transportation, Construction building technology, and Robotics.

## 1. Introduction

Optimization, referred to as finding the best solution to a problem, is desirable in many fields such as engineering, operations research, economics, supply chain, computer science, chemistry, medicine, physics, etc. (Coelho, Ayala & Mariani 2014; Ng & Li 2014; Wang et al. 2013).

Finding an optimal/sub-optimal solutions to an optimization problem is very challenging, especially for complex large-scale problems. Generally speaking, there are two main optimization solution methods, namely deterministic methods and stochastic methods (Hanagandi & Nikolaou 1998). Both methods have advantages and disadvantages.

Deterministic methods can guarantee the optimal solutions for some problems which have a certain features. However, deterministic methods might fail when they deal with black-box problems and/or complex large-scale problems, due to the issue of combinatorial explosion. Stochastic methods are capable of working with any kind of problems but they have a weak capability to guarantee the optimal solutions (Liberti & Kucherenko 2005; Moles, Mendes & Banga 2003). Nevertheless, no method could find the global optimal solution to a general optimization problem in finite computing time (Boender & Romeijn 1995).

According to Dao, Abhary & Marian (2016), stochastic methods are more popular than deterministic methods for solving complex large-scale problems. The reasons behind the popularity of the stochastic methods could be as follows. First, stochastic methods do not require an advanced mathematical knowledge. Second, stochastic approaches can easily handle many complex constraints. Finally, the solution quality provided by stochastic methods has been significantly improved due to the advancement of computing techniques/machines.

To date, there have been a number of stochastic methods developed to solve many complex large-scale optimization problems in various fields such as Genetic Algorithm, Cuckoo Search, Tabu Search, Simulated Annealing, Particle Swarm Optimization, Ant Colony Optimization, Hill Climbing, Harmony Search, Greedy Algorithm, Pattern Search, Stochastic Tunneling, Differential Evolution, and Cross-entropy Method. Each method has its own strengths and weaknesses. To help researchers and practitioners select the best algorithm to solve the real-life problems in the fields of Engineering, Computer science, Operations research, Mathematics, Physics, Chemistry, Automation control systems, Materials science, Energy fuels, Mechanics, Telecommunications, Thermodynamics, Optics, Environmental sciences ecology, Water resources, Transportation, Construction building technology, and Robotics, a quantitative analysis of the popularity of top 14 stochastic optimization algorithms in the last ten years from

2007 to 2017 is provided herein. This quantitative analysis of the popularity of the algorithms is based on the number of journal articles and citations.

## 2. Data Collection

Web of Science (WoS), originally produced by the Institute for Scientific Information (ISI) and currently maintained by Clarivate Analytics, is the world's leading citation database, with multidisciplinary information from over 18,000 high impact journals, over 180,000 conference proceedings, and over 80,000 books from around the world (Clarivate.Analytics 2017). All journals indexed by WoS are the prestigious peer-reviewed journals which have met the high standards of an objective evaluation process (Thomson.Reuters 2016). In this note, journal articles in WoS database are used to analyze the popularity of stochastic optimization algorithms in the last ten years from 2007 to 2017.

When searching for journal articles in WoS database, we used 14 key words corresponding to 14 stochastic optimization algorithms, i.e. "Artificial bee colony", "Ant colony optimization", "Cuckoo search", "Cross-entropy method", "Differential evolution", "Genetic algorithm", "Greedy algorithm", "Hill-climbing", "Harmony search", "Particle swarm optimization", "Pattern search", "Simulated annealing", "Stochastic tunneling", and "Tabu search". The key words were used one by one, and the searching procedure is shown in Fig. 1. In addition, the search results obtained using the procedure in Fig. 1 will be narrowed down to journal articles only, and then classified based on the fields such as Engineering, Computer science, Operations research, Mathematics, etc. The final results, i.e. the number of journal articles and citations, will be shown and discussed in the next Section.

Fig. 1: Data collection procedure

## 3. Result and Discussion

The number of journal articles using 14 stochastic optimization algorithms (Artificial bee colony, Ant colony optimization, Cuckoo search, Cross-entropy method, Differential evolution, Genetic algorithm, Greedy algorithm, Hill-climbing, Harmony search, Particle swarm optimization, Pattern search, Simulated annealing, Stochastic tunneling, and Tabu search) in 18 different research fields (Engineering, Computer science, Operations research, Mathematics, Physics, Chemistry, Automation control systems, Materials science, Energy fuels, Mechanics, Telecommunications, Thermodynamics, Optics, Environmental sciences ecology, Water resources, Transportation, Construction building technology, and Robotics) during a ten year period from 2007 to 2017 is shown in Table 1 and Fig. 2.

Table 1: The number of journal articles using stochastic optimization algorithms in different fields during a period of 2007-2017

| No. | Algorithms | Engineering | Computer science | Operations research | Mathematics | Physics | Chemistry | Automation control systems | Materials science | Energy fuels | Mechanics | Telecom-munications | Thermo-dynamics | Optics | Environmental sciences ecology | Water resources | Transport-ation | Construction building technology | Robotics |
|---|---|---|---|---|---|---|---|---|---|---|---|---|---|---|---|---|---|---|---|
| 1 | Artificial bee colony | 390 | 398 | 79 | 83 | 31 | 22 | 58 | 16 | 32 | 26 | 41 | 22 | 12 | 6 | 6 | 5 | 9 | 3 |
| 2 | Ant colony optimization | 432 | 504 | 128 | 98 | 20 | 34 | 64 | 18 | 19 | 20 | 70 | 13 | 18 | 22 | 17 | 17 | 13 | 17 |
| 3 | Cuckoo search | 161 | 138 | 19 | 10 | 11 | 8 | 20 | 5 | 19 | 10 | 14 | 11 | 3 | 6 | 2 | 0 | 6 | 0 |
| 4 | Cross-entropy method | 35 | 19 | 13 | 16 | 9 | 0 | 0 | 0 | 0 | 2 | 12 | 0 | 4 | 0 | 0 | 0 | 1 | 0 |
| 5 | Differential evolution | 931 | 862 | 195 | 201 | 83 | 48 | 128 | 47 | 96 | 58 | 93 | 45 | 27 | 16 | 33 | 11 | 20 | 12 |
| 6 | Genetic algorithm | **4017** | **2395** | **769** | **618** | **460** | **449** | **429** | **400** | **382** | **369** | **294** | **254** | **228** | **189** | **188** | **116** | **111** | **57** |
| 7 | Greedy algorithm | 50 | 61 | 29 | 65 | 7 | 2 | 5 | 3 | 6 | 3 | 9 | 4 | 0 | 2 | 0 | 2 | 0 | 1 |
| 8 | Hill-climbing | 27 | 29 | 7 | 7 | 3 | 1 | 5 | 3 | 5 | 3 | 3 | 1 | 3 | 1 | 0 | 1 | 1 | 3 |
| 9 | Harmony search | 310 | 238 | 65 | 71 | 13 | 7 | 37 | 9 | 30 | 21 | 7 | 11 | 4 | 8 | 17 | 2 | 12 | 3 |
| 10 | Particle swarm optimization | 2070 | 1720 | 311 | 447 | 236 | 126 | 300 | 95 | 236 | 175 | 198 | 133 | 110 | 61 | 66 | 15 | 27 | 49 |
| 11 | Pattern search | 66 | 42 | 14 | 27 | 4 | 3 | 8 | 1 | 7 | 8 | 5 | 4 | 3 | 4 | 1 | 1 | 2 | 0 |
| 12 | Simulated annealing | 628 | 413 | 202 | 174 | 110 | 63 | 78 | 41 | 34 | 52 | 37 | 20 | 26 | 23 | 23 | 18 | 12 | 7 |
| 13 | Stochastic tunneling | 1 | 2 | 0 | 1 | 2 | 0 | 0 | 0 | 0 | 0 | 1 | 0 | 0 | 0 | 0 | 0 | 0 | 0 |
| 14 | Tabu search | 323 | 265 | 267 | 72 | 21 | 9 | 35 | 5 | 9 | 6 | 30 | 2 | 7 | 2 | 7 | 25 | 3 | 3 |

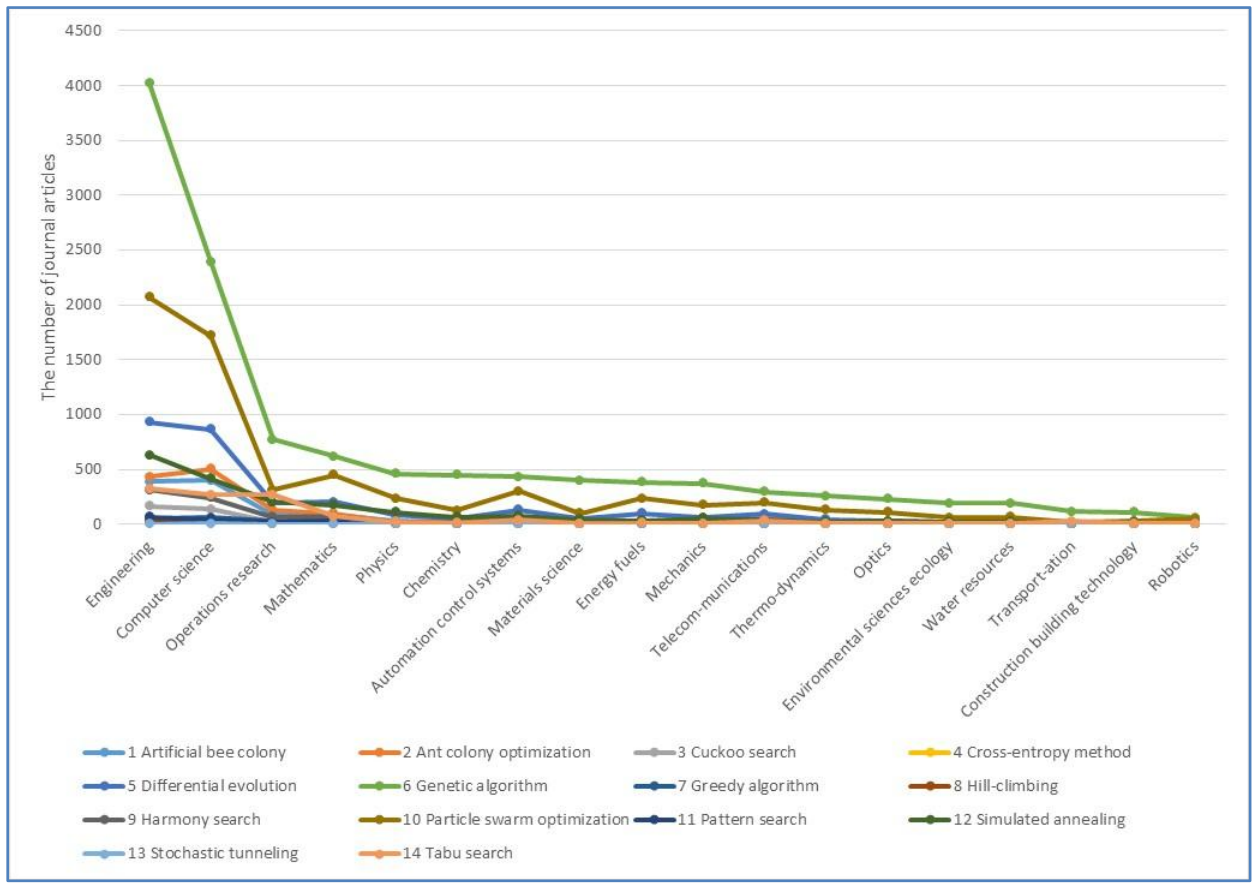

Fig. 2: Visualization of the popularity of stochastic optimization algorithms in different fields in ten years from 2007-2017

As can be seen from Table 1 and Fig. 2, Genetic algorithm is the most popular stochastic optimization algorithm in all 18 different research fields in the last ten years from 2007 to 2017. More interestingly, there are a significant number of journal articles using Genetic algorithm in Engineering (4017), Computer science (2395), Operations research (769), Mathematics (618), and Physics (460). The second most popular stochastic optimization algorithm used in Engineering, Computer science, Operations research, Mathematics, Physics, Chemistry, Automation control systems, Material science, Energy fuels, Mechanics, Telecommunications, Thermodynamics, Optics, Environmental sciences ecology, Water sciences, Construction building technology, and Robotics is Particle swarm optimization algorithm. It should be noted that in the field of Transportation, the second most popular stochastic optimization algorithm is Simulated annealing.

The top three research fields, in which the stochastic optimization algorithms are frequently used, are Engineering, Computer science, and Operations research, as indicated in Table 1 and Fig. 2. Figs. 3-5 show the publication trends of the algorithms over the years in the top three research fields. Obviously, Genetic algorithm is the most popular algorithm in all top three research fields in every year from 2007 to 2017. Generally speaking, the publication trends of the algorithms in the fields of Engineering and Computer science have been increasing over the years; while those trends in Operations research increased to reach a peak at the period of 2009 to 2012, and then went down after that.

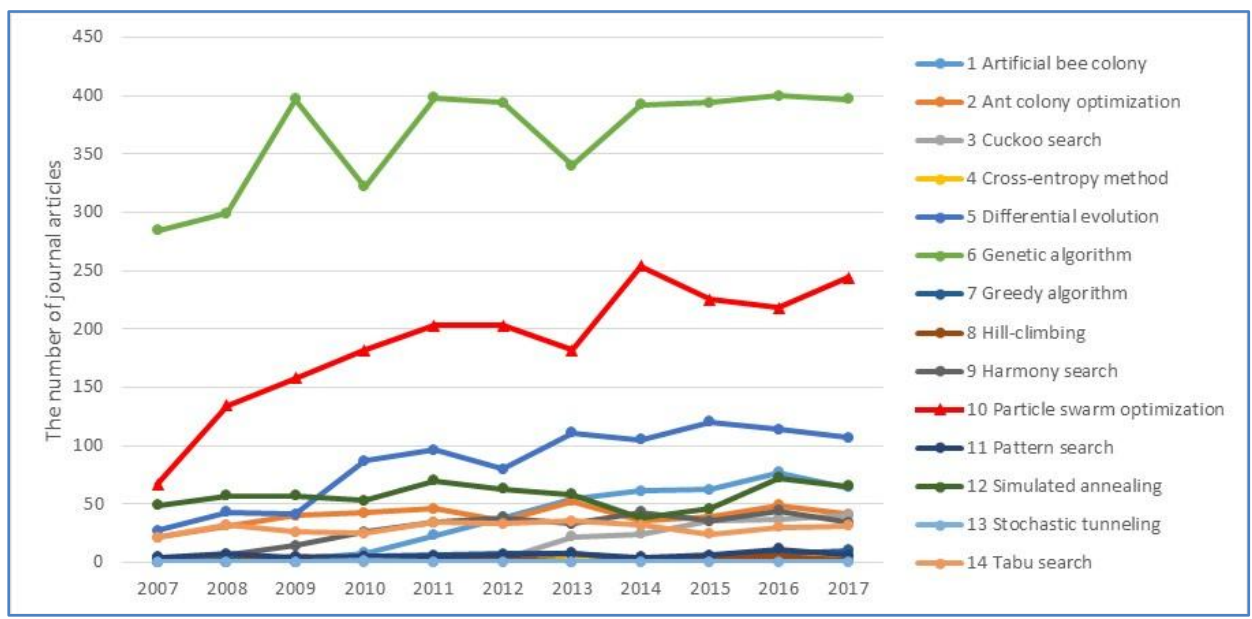

Fig. 3: Publication trends of the algorithms over the years in the field of engineering

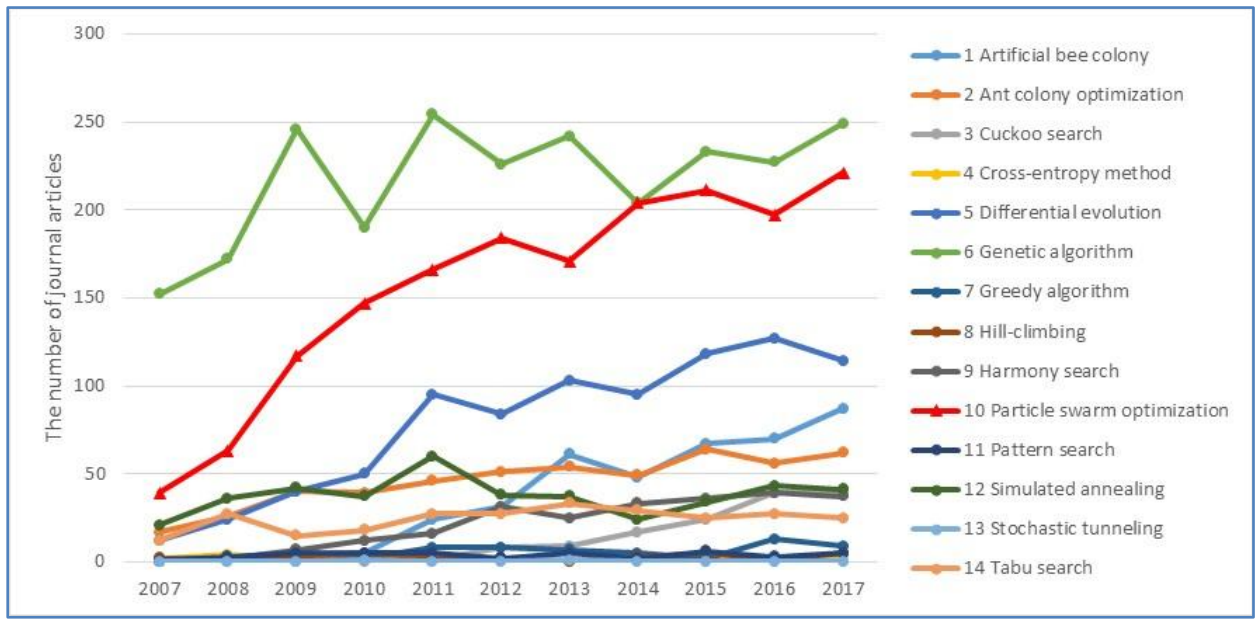

Fig. 4: Publication trends of the algorithms over the years in the field of computer science

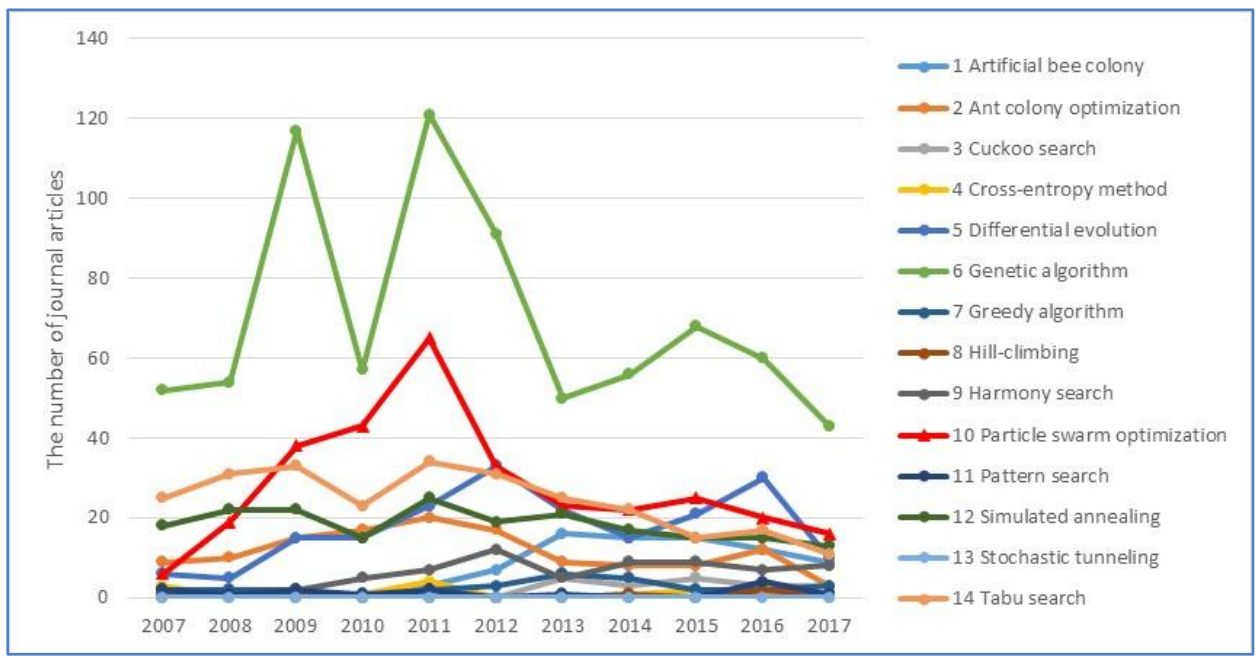

Fig. 5: Publication trends of the algorithms over the years in the field of operations research

Beside the number of journal articles, the popularity of the stochastic optimization algorithms is also measured in terms of the number of citations. Table 2 shows the number of citations of the journal articles using the stochastic optimization algorithms in different fields during the ten year period from 2007 to 2017. It can be seen from Table 2 that Genetic algorithm is the most highly cited in the fields of Engineering (47572), Operations research (14782), Chemistry (4860), Materials science (3688), Energy fuels (6862), Mechanics (5017), Telecommunications (2553), Thermodynamics (3671), Optics (1748), Environmental sciences ecology (2213), Water resources (2484), Transportation (1242), and Construction building technology (1938); while Particle swarm optimization is the most highly cited in the fields of Computer science (33535), Mathematics (6934), Physics (3868), Automation control systems (7300), and Robotics (442).

Table 2: The number of citations of journal articles using stochastic optimization algorithms in different fields during a period of 2007-2017

| No. | Algorithms | Engineering | Computer science | Operations research | Mathematics | Physics | Chemistry | Automation control systems | Materials science | Energy fuels | Mechanics | Telecom-munications | Thermo-dynamics | Optics | Environmental sciences ecology | Water resources | Transport-ation | Construction building technology | Robotics |
|---|---|---|---|---|---|---|---|---|---|---|---|---|---|---|---|---|---|---|---|
| 1 | Artificial bee colony | 5921 | 10917 | 3826 | 4621 | 482 | 230 | 1306 | 74 | 742 | 603 | 422 | 490 | 48 | 41 | 73 | 109 | 80 | 12 |
| 2 | Ant colony optimization | 7197 | 8529 | 3406 | 977 | 111 | 401 | 966 | 233 | 450 | 355 | 582 | 201 | 124 | 346 | 395 | 584 | 511 | 138 |
| 3 | Cuckoo search | 2572 | 2403 | 618 | 364 | 321 | 61 | 320 | 96 | 413 | 117 | 72 | 227 | 15 | 34 | 20 | 0 | 107 | 0 |
| 4 | Cross-entropy method | 281 | 209 | 184 | 105 | 28 | 0 | 0 | 0 | 0 | 3 | 89 | 0 | 26 | 0 | 0 | 0 | 7 | 0 |
| 5 | Differential evolution | 14159 | 22786 | 3724 | 2697 | 1304 | 840 | 2646 | 354 | 2218 | 1650 | 1196 | 1163 | 258 | 205 | 477 | 143 | 279 | 61 |
| 6 | Genetic algorithm | **47572** | 31242 | **14782** | 5348 | 3850 | **4860** | 4995 | **3688** | **6862** | 5017 | 2553 | **3671** | 1748 | **2213** | 2484 | **1242** | **1938** | 378 |
| 7 | Greedy algorithm | 695 | 585 | 797 | 406 | 171 | 12 | 73 | 15 | 87 | 53 | 25 | 63 | 0 | 18 | 0 | 11 | 0 | 30 |
| 8 | Hill-climbing | 565 | 304 | 51 | 14 | 14 | 6 | 30 | 87 | 131 | 9 | 22 | 5 | 10 | 46 | 0 | 11 | 5 | 17 |
| 9 | Harmony search | 5859 | 3918 | 1344 | 2443 | 194 | 79 | 707 | 146 | 813 | 841 | 80 | 402 | 35 | 200 | 344 | 14 | 391 | 32 |
| 10 | Particle swarm optimization | 34297 | **33535** | 8235 | **6934** | 3868 | 1275 | **7300** | 1027 | 5296 | 3403 | 2115 | 3081 | 907 | 712 | 974 | 275 | 476 | **442** |
| 11 | Pattern search | 745 | 529 | 255 | 439 | 90 | 34 | 36 | 0 | 182 | 110 | 67 | 61 | 19 | 43 | 22 | 13 | 12 | 0 |
| 12 | Simulated annealing | 7605 | 6575 | 3394 | 1507 | 947 | 430 | 835 | 438 | 706 | 530 | 227 | 200 | 124 | 185 | 199 | 189 | 255 | 146 |
| 13 | Stochastic tunneling | 4 | 11 | 0 | 6 | 18 | 0 | 0 | 0 | 0 | 0 | 7 | 0 | 0 | 0 | 0 | 0 | 0 | 0 |
| 14 | Tabu search | 4549 | 3544 | 5120 | 532 | 147 | 94 | 559 | 49 | 174 | 160 | 294 | 108 | 58 | 7 | 39 | 688 | 62 | 23 |

## 4. Conclusion

In this note, Web of Science, the world's leading citation database, was used to determine the popularity of 14 stochastic optimization algorithms in the last ten years from 2007 to 2017 in 18 different research fields including Engineering, Computer science, Operations research, Mathematics, Physics, Chemistry, Automation control systems, Materials science, Energy fuels, Mechanics, Telecommunications, Thermodynamics, Optics, Environmental sciences ecology, Water resources, Transportation, Construction building technology, and Robotics. The data showed that Engineering, Computer science, and Operations research are the top three research fields in which the stochastic optimization algorithms are the most frequently used. In terms of the number of journal articles, Genetic algorithm is the most popular stochastic optimization algorithm in all 18 different research fields, followed by Particle swarm optimization algorithm. In terms of citations, Genetic algorithm is the most highly cited in 13 research fields; while Particle swarm optimization algorithm is the most highly cited in 5 research fields.

## References

Boender, CGE & Romeijn, HE 1995, 'Stochastic methods', in R Horst & PM Pardalos (eds), *Handbook of Global Optimization*, Kluwer Academic Publishers, Boston

Clarivate.Analytics 2017, *Web of Science Core Collection*, viewed 10 December 2017, <https://clarivate.com/products/web-of-science/databases/>.

Coelho, LDS, Ayala, HVH & Mariani, VC 2014, 'A self-adaptive chaotic differential evolution algorithm using gamma distribution for unconstrained global optimization', *Applied Mathematics and Computation,* vol. 234, pp. 452-459.

Dao, SD, Abhary, K & Marian, R 2016, 'An improved structure of genetic algorithms for global optimisation', *Progress in Artificial Intelligence,* vol. 5, no. 3, pp. 155-163.

Hanagandi, V & Nikolaou, M 1998, 'A hybrid approach to global optimization using a clustering algorithm in a genetic search framework', *Computers & Chemical Engineering,* vol. 22, no. 12, pp. 1913-1925.

Liberti, L & Kucherenko, S 2005, 'Comparison of deterministic and stochastic approaches to global optimization', *International Transactions in Operational Research,* vol. 12, no. 3, pp. 263-285.

Moles, CG, Mendes, P & Banga, JR 2003, 'Parameter estimation in biochemical pathways: a comparison of global optimization methods', *Genome Research,* vol. 13, no. 11, pp. 2467–2474.

Ng, CK & Li, D 2014, 'Test problem generator for unconstrained global optimization', *Computers & Operations Research,* vol. 51, no. 0, pp. 338-349.

Thomson.Reuters 2016, *Science citation index expanded*, Thomson Reuters, viewed 16 November 2016, <http://thomsonreuters.com/science-citation-index-expanded/?subsector=scholarly-search-and-discovery>.

Wang, Y, Huang, J, Dong, WS, Yan, JC, Tian, CH, Li, M & Mo, WT 2013, 'Two-stage based ensemble optimization framework for large-scale global optimization', *European Journal of Operational Research,* vol. 228, no. 2, pp. 308-320.